Research and Applications

# Multimodal integration of longitudinal noninvasive diagnostics for survival prediction in immunotherapy using deep learning

Melda Yeghaian, MSc[1,2,3,*], Zuhir Bodalal, MD, MSc[2,3], Daan van den Broek, PhD[4], John B.A.G. Haanen, MD, PhD[5,6,7,8], Regina G.H. Beets-Tan, MD, PhD[2,3,9], Stefano Trebeschi, PhD[2,3], Marcel A.J. van Gerven, PhD[1]

[1]Department of Machine Learning and Neural Computing, Donders Institute for Brain, Cognition and Behaviour, Radboud University, Nijmegen 6525 GD, The Netherlands, [2]Department of Radiology, The Netherlands Cancer Institute, Amsterdam 1066 CX, The Netherlands, [3]GROW School for Oncology and Developmental Biology, Maastricht University, Maastricht 6229 ER, The Netherlands, [4]Department of Laboratory Medicine, The Netherlands Cancer Institute, Amsterdam 1066 CX, The Netherlands, [5]Department of Medical Oncology, The Netherlands Cancer Institute, Amsterdam 1066 CX, The Netherlands, [6]Division of Molecular Oncology and Immunology, Oncode Institute, Amsterdam 1066 CX, The Netherlands, [7]Department of Medical Oncology, Leiden University Medical Center, Leiden 2333 ZG, The Netherlands, [8]Melanoma Clinic, Centre Hospitalier Universitaire Vaudois, Lausanne 1005, Switzerland, [9]Faculty of Health Science, University of Southern Denmark, Odense 5230, Denmark

*Corresponding author: Melda Yeghaian, MSc, Department of Machine Learning and Neural Computing, Donders Institute for Brain, Cognition and Behaviour, Radboud University, Thomas van Aquinostraat 4, 6525 GD Nijmegen, Gelderland, The Netherlands (melda.yeghaian@donders.ru.nl)

S. Trebeschi and M.A.J. van Gerven contributed equally to this work.

## Abstract

**Objectives:** Immunotherapies have revolutionized the landscape of cancer treatments. However, our understanding of response patterns in advanced cancers treated with immunotherapy remains limited. By leveraging routinely collected noninvasive longitudinal and multimodal data with artificial intelligence, we could unlock the potential to transform immunotherapy for cancer patients, paving the way for personalized treatment approaches.

**Materials and Methods:** In this study, we developed a novel artificial neural network architecture, multimodal transformer-based simple temporal attention (MMTSimTA) network, building upon a combination of recent successful developments. We integrated pre- and on-treatment blood measurements, prescribed medications, and CT-based volumes of organs from a large pan-cancer cohort of 694 patients treated with immunotherapy to predict mortality at 3, 6, 9, and 12 months. Different variants of our extended MMTSimTA network were implemented and compared to baseline methods, incorporating intermediate and late fusion-based integration methods.

**Results:** The strongest prognostic performance was demonstrated using a variant of the MMTSimTA model with area under the curves of 0.84 ± 0.04, 0.83 ± 0.02, 0.82 ± 0.02, 0.81 ± 0.03 for 3-, 6-, 9-, and 12-month survival prediction, respectively.

**Discussion:** Our findings show that integrating noninvasive longitudinal data using our novel architecture yields an improved multimodal prognostic performance, especially in short-term survival prediction.

**Conclusion:** Our study demonstrates that multimodal longitudinal integration of noninvasive data using deep learning may offer a promising approach for personalized prognostication in immunotherapy-treated cancer patients.

## Introduction

During cancer treatment, noninvasive data, such as laboratory blood test results and radiological imaging, are routinely collected by clinicians to guide clinical decision-making. Noninvasive data offer safe and cost-effective observables that can be frequently collected, reducing the need for repeated invasive procedures. Moreover, various sources of medical data are also routinely monitored by clinicians during treatment, since each modality measures a different aspect and has a different, sometimes complementary, ability to quantify the health status of patients. Combining these modalities using artificial intelligence (AI) could potentially lead to better patient stratification strategies and provide individualized care.[1,2] In addition, patient data are routinely collected longitudinally as diseases and treatments evolve over time. Therefore, capturing the temporal dimension, as well, by analyzing longitudinal patient data using AI is key for the prediction of patient outcomes (eg, early response to treatment and prognosis), offering personalized treatment strategies.[3,4]

Immunotherapies such as immune checkpoint inhibitors (ICI) have changed the landscape of cancer care. While they have transformed 20% of fatal cancers into manageable chronic conditions, providing patients with better survival and quality of life, the remaining 80% of all cancers show poor response to therapy.[4] Immunotherapy is also costly and can cause adverse events. A number of clinical biomarkers, such as the tumor mutation burden, programmed death ligand 1 expression and DNA repair deficiency (eg, microsatellite instability and DNA polymerase $\varepsilon$ (POLE) mutation), are clinically used to predict response to ICIs in certain cancers.[4,5] However, the evolving nature of the tumor microenvironment and tumor characteristics over time changes the sensitivity to ICIs, limiting durable response patterns in advanced cancers.[4] With the infeasibility of obtaining multiple tumor biopsies over time, and the challenges in continuous monitoring via circulating tumor DNA sequencing, there is a need to develop noninvasive longitudinal and multimodal biomarkers using large-scale databases and AI to predict response and efficacy of immunotherapy.[4,5]

In a multimodal setting, several studies have emerged for response, progression, and survival prediction in immunotherapy using data at a single time point.[6–16] Cancer prognostication using nonlongitudinal multimodal data in other therapies was also extensively studied in the literature.[17–30] Considering the need to analyze data longitudinally in cancer immunotherapy, several longitudinal studies using serial medical examinations have addressed this demand in recent years.[31–42] Among these, only a few employed multimodal data in addition to longitudinal (serial) imaging data.[31,33,39,40]

Yang et al[31] introduced simple temporal attention (SimTA), an attention module specific for asynchronous time series (longitudinal) analysis in healthcare, showing its successful application in multimodal and longitudinal settings for response prediction of non-small cell lung cancer (NSCLC) patients treated with immunotherapy.[31,33] The latter studies employing SimTA provided valuable insights, though they did not extensively explore the experimental aspects given the simplicity of SimTA, which is designed to learn the temporal relations of data linearly. Transformers have shown widespread success in multimodal learning and time series analysis in recent years,[43–46] becoming the default choice of architecture in various domains. We hypothesize that embedding the SimTA module within a transformer encoder, with additional non-linearity and skip connection, could further improve the performance of the models for longitudinal and multimodal data analysis.

In this study, we developed a novel artificial neural network architecture by extending the transformer encoder architecture[43] with the longitudinal attention module[31] for overall survival prediction in a large pan-cancer cohort treated with immunotherapy. Specifically, we used routinely collected noninvasive pretreatment and on-treatment longitudinal and multimodal data, consisting of laboratory-based blood measurements, body organ volumes from CT imaging, and prescribed medications information. We predicted mortality at 3, 6, 9, and 12 months in an end-to-end multimodal multitask classification algorithm.

## Materials and methods
### Study cohort and ethical approval
Institutional review board (IRB) approval was obtained from the Netherlands Cancer Institute (approval number: IRBd23-219), where the need for project-specific informed consent was waived. We retrospectively collected all cancer patients treated with ICIs (specifically nivolumab, pembrolizumab, ipilimumab, or a combination of nivolumab and ipilimumab) in a non-neoadjuvant setting at the Netherlands Cancer Institute between January 2016 and August 2022. We only included patients with a single cancer diagnosis who received a single line of immunotherapy with the abovementioned ICIs. Patients lost to follow-up, those with a follow-up duration of <3 months from the start of treatment, and those without a 12-month prediction window after the last included follow-up examination were excluded.

For each included patient, we collected noninvasive multimodal longitudinal data, including laboratory blood test measurements, contrast-enhanced CT imaging, and prescribed medications. Blood data consisted of the measurements of routine blood markers, while imaging data included the volumes of body organs (in $mm^3$) extracted from thoracic and abdominal CT scans. Four categories of drugs, namely, anticancer therapies, immunosuppressants, opioids, and corticosteroids, were also included. We included data from each modality acquired between 3 months pretreatment and up to 1 year after. Death dates of patients were also retrieved, and survival prediction was approached as a multitask binary classification problem (eg, survival at 3, 6, 9, and 12 months).

### Data preprocessing
From the included longitudinal thoracic and abdominal contrast-enhanced CT scans, we automatically segmented 117 anatomical structures and calculated their volumes using the TotalSegmentator model.[47] Certain volumes of interest were occasionally absent in the scan due to a limited field of view on the image, medical reasons such as surgical resection of the structure, or technical reasons such as absent/incomplete segmentation by the algorithm. Structures with ≥50% missing values were dropped, resulting in a final set of 100 volume features per scan. We excluded blood tests with >60% missing values, which resulted in the retention of 33 blood markers. Multivariate iterative imputation was used to fill in missing values of blood markers.[48] Imaging and blood features were scaled using statistics that are robust to outliers (median and interquartile range).[48] Imputation and scaling were performed based on the training sets. Anatomical Therapeutic Chemical classification codes of medications were tokenized and encoded using PyTorch embedding layers.[49] More details on the features utilized are provided in the data preprocessing section of the Supplementary Material and Table S1.

### Model design and architecture
We developed a novel architecture by extending the transformer encoder block with the SimTA module proposed by Yang et al[31] for encoding longitudinal patient data. The modified architecture, the transformer-based SimTA (TSimTA) network, is not only suitable for longitudinal data but it also substitutes the expensive self-attention (SA) layer of transformers with the SimTA module, which linearly encodes time intervals. This is advantageous for handling longitudinal patient data, improving scalability to long sequences.

The SimTA module was originally proposed in a multimodal longitudinal setting,[31] which we refer to as MMSimTA. To adapt the extended transformer-based architecture for multimodality, we also followed a common intermediate fusion approach, which trains a multilayer perceptron (MLP) after concatenating the representations of each individual modality, obtained through modality-specific TSimTA blocks, into the same representation space, allowing joint training of the MMTSimTA multimodal network,[50] similar to MMSimTA. We also investigated a multimodal model variant that included a transformer block with multi-head SA between the concatenated features and the MLP. Different variants of the baseline multimodal MMSimTA[31] and MMTSimTA models were also implemented for comparison as detailed in Section S1.2.

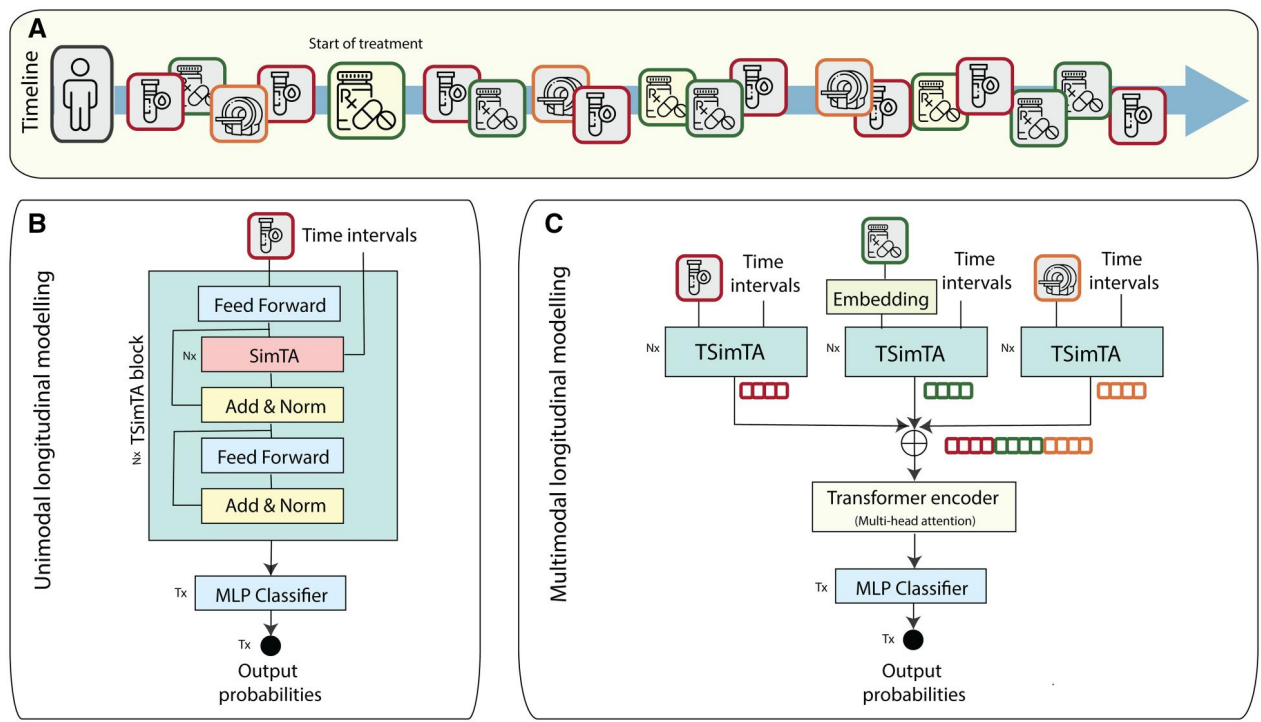

**Figure 1.** A schematic overview of the extended transformer-based models. (A) The patient timeline with non-uniformly acquired noninvasive longitudinal and multimodal (blood, medication, and imaging) examinations. (B) The extended unimodal model TSimTA. (C) The extended transformer-based multimodal model MMTSimTA. TSimTA blocks were replaced with SimTA blocks[31] in all baseline experiments. Abbreviations: T: the number of prediction tasks; N: the number of SimTA and/or TSimTA blocks.

In addition to the intermediate fusion approach, we implemented a late fusion model-agnostic strategy, also used in Yeghaian et al,[40] for comparison. This strategy averages the unimodal probabilities of individual modalities, produced by both baseline and extended transformer-based models, SimTA and TSimTA, into a unified and integrated decision, referred to as MA-MMSimTA and MA-MMTSimTA, respectively. Figure 1 depicts the architecture of the extended unimodal and multimodal model(s).

### Model training and validation strategies

The SimTA module was originally built with the assumption that the most recent examinations are the most important ones during the patient timeline[33]; therefore, we trained the models including data from all existing examinations up to randomly selected follow-up (FU) durations (between 3 months after the start of treatment [SoT] up to 1 year).

The models were trained to simultaneously predict the overall survival of patients at different prediction endpoints, namely, at 3, 6, 9, and 12 months from the (randomly) selected FU duration. All modalities and prediction tasks were jointly trained in an end-to-end manner with the preprocessed features of arbitrary numbers of available modalities using multimodal dropout.[51] We adopted a 3-fold stratified cross-validation approach for evaluating the trained models. Validation was performed using early on-treatment data (up to 3 months after SoT) and, occasionally, late on-treatment (up to 6 months after SoT) data. More details on model training and validation are provided in the supplements. Our code extended the publicly available SimTA unimodal module, and will be publicly shared.

### Statistical analysis

The area under the receiver operating characteristic curve (AUC) was used to evaluate the predictive performance of the studied models. In each cross-validation fold, the statistical significance of the classifiers was tested using the Mann-Whitney $U$ test. DeLong tests assessed the statistical significance of AUC differences between classifiers in each fold, comparing their performance. The $P$-values resulting from the DeLong test in each fold were combined using Fisher's method. Mean AUCs and their standard deviation across the 3 cross-validation folds were reported. A $P$-value $<.05$ was considered statistically significant. All analyses were performed using Python (v3.10.13) and the following packages: PyTorch 2.0.1, Scikit-learn 1.3.0, TotalSegmentator 2.0.5, NumPy 1.25.2, and Pandas 2.0.3.

## Results

### Patient characteristics

We identified 1129 cancer patients treated with a single line of immunotherapy using nivolumab, pembrolizumab, ipilimumab, or a combination of nivolumab and ipilimumab in a non-neoadjuvant setting. Among them, 9 were lost to FU patients and 146 had multiple (primary) tumors and were, therefore, excluded. A total of 280 additional patients with insufficient longitudinal information were excluded. Those patients either had <3-month FU data or longitudinal information <1 year after a 3-month FU (Table S2). This exclusion was performed to ensure a minimum of 3 months of longitudinal FU duration in the study, allowing for the evaluation of our longitudinal method. The final patient cohort used in our analyses consisted of a total of 694 cancer patients, with 11 different tumor types, and at least 3-month FU data. Table 1 and Figure S1 offer more detailed patient characteristics for the included patient cohort.

**Table 1.** Patient characteristics.

| Characteristics | Values |
| --- | --- |
| Age (years), median (interquartile range) | 63 (56-70) |
| Sex, n (%) | |
|   Male | 363 (52%) |
|   Female | 331 (48%) |
| Malignancy class, n (%) | |
|   Respiratory and intrathoracic organs | 357 (51%) |
|   Melanoma and other skin malignancies | 170 (24%) |
|   Urinary tract | 82 (12%) |
|   Female genital organs | 26 (4%) |
|   Lip, oral cavity and pharynx | 23 (3%) |
|   Digestive organs | 20 (3%) |
|   Breast | 7 (1%) |
|   Male genital organs | 6 (1%) |
|   Stated or presumed to be primary, of lymphoid, hematopoietic and related tissue | 2 (0.3%) |
|   Thyroid and other endocrine glands | 1 (0.1%) |
| Treatment, n (%) | |
|   Nivolumab | 344 (50%) |
|   Pembrolizumab | 267 (38%) |
|   Ipilimumab | 67 (10%) |
|   Nivolumab and Ipilimumab | 16 (2%) |
| Outcome, n (%) | |
|   Death | 537 (77.38%) |
|   Alive | 157 (22.62%) |
| Longest follow-up time for 12-month survival prediction (days), median (interquartile range) | 418 (216-1334) |

Not all diagnostic data were available for our included patient cohort: 693 patients had recorded routine blood test results, 402 patients had CT imaging, and all patients had prescribed medication information. A total of 11 249 blood tests, 14 849 medication records, and 1337 CT scans were retrieved.

### Extended transformer-based architecture demonstrates equal unimodal prognostic performance

We ran unimodal data-based evaluation using the baseline SimTA-based and the extended TSimTA-based models, including data acquired between 3 months pretreatment and 3 months after, to compare their performance for survival prediction. The extended transformer-based model, TSimTA, did not always show improved performance across all 4 prediction endpoints (AUCs$_{SimTA}$ of 0.83, 0.80, 0.80, 0.80 vs AUCs$_{TSimTA}$ of 0.82, 0.81, 0.82, 0.81 for the prediction of 3-, 6-, 9-, and 12-month survival in the best-performing individual modality, blood data). The 2 unimodal models did not statistically differ in terms of the AUCs of the predictive performance (P-values >.05 in all the aforementioned endpoints). The numbers of patients in the test sets are provided in Table S3. The performance of the remaining unimodal modalities is detailed in Table 2 and Table S4. The correlations of the unimodal probabilities produced by the modality-specific models are provided in Table S5.

### Extended transformer-based architecture shows improved multimodal prognostic performance

Several experiments were conducted with different variants of multimodal baseline and extended transformer-based models to compare the prognostic performance, including data acquired between 3 months pretreatment and 3 months after. The best variant of the baseline multimodal SimTA models, MMSimTA, showed equal or decreased prognostic performance compared to the prognostic performance of the best-performing unimodal modality, blood data (AUCs$_{MMSimTA}$ of 0.83, 0.80, 0.78, 0.78 vs AUCs$_{SimTA}$ of 0.83, 0.80, 0.80, 0.80; DeLong P-values >.05 for the prediction of 3-, 6-, 9-, and 12-month survival, respectively; Figure S2). In comparison, the best variant of the extended transformer-based multimodal TSimTA model, MMTSimTA, performed significantly better than the best variant of the baseline MMSimTA (AUCs$_{MMTSimTA}$ of 0.84, 0.83, 0.82, 0.81 vs AUCs$_{MMSimTA}$ of 0.83, 0.80, 0.78, 0.78; DeLong P-values of >.05, .01, .001, .01 for the prediction of 3-, 6-, 9- and 12-month survival, respectively). Comparing MMTSimTA with the best-performing unimodal TSimTA model, multimodal performance showed higher AUCs in the earlier prediction endpoints, while showing equal performance in later prediction endpoints, (AUCs$_{MMTSimTA}$ of 0.84, 0.83, 0.82, 0.81 vs AUCs$_{TSimTA}$ of 0.82, 0.81, 0.82, 0.81; DeLong P-values of >.05, .03, >.05, >.05 for 3-, 6-, 9-, and 12-month survival prediction, respectively; Figure 2A). Table 3 summarizes the results of the multimodal experiments. Additional results are provided in the Supplementary Results Section as well as in Tables S6 and S7.

### Extended transformer-based architecture improves multimodal prognostic performance with intermediate over late fusion

The jointly trained best-performing baseline and extended multimodal transformer-based models were compared against a **m**odel-**a**gnostic late fusion integration method, MA-MMSimTA, MA-MMTSimTA, respectively. MA-MMSimTA with late fusion performed better than the baseline MMSimTA incorporating intermediate fusion (AUCs$_{MA-MMSimTA}$ of 0.85, 0.81, 0.81, 0.81 vs AUCs$_{MMSimTA}$ of 0.83, 0.80, 0.78, 0.78; DeLong P-values of >.05, >.05, .03, .03 for the prediction of 3-, 6-, 9-, and 12-month survival, respectively). In contrast, the extended transformer-based model, MMTSimTA, incorporating intermediate fusion, was better able to predict overall survival than MA-MMTSimTA, incorporating late fusion, in almost all prediction endpoints (AUCs$_{MMTSimTA}$ of 0.84, 0.83, 0.82, 0.81 vs AUCs$_{MA-MMTSimTA}$ of 0.84, 0.80, 0.80, 0.79; DeLong P-values of .005, .02, .04, >.05 for 3-, 6-, 9-, and 12-month survival prediction, respectively). Comparing MA-MMSimTA to the best baseline unimodal SimTA model of blood data, we observed that integrating modalities via late fusion performed better than the best unimodal modality (AUCs$_{MA-MMSimTA}$ of 0.85, 0.81, 0.81, 0.81 vs AUCs$_{SimTA}$ of 0.83, 0.80, 0.80, 0.80; DeLong P-values of >.05 for the prediction of 3-, 6-, 9-, and

**Table 2.** Comparative analysis of the performance of the unimodal classifiers, SimTA and TSimTA, using up to 3-month on-treatment data (blood markers, organ volumes, and medications).

| | | Mean AUC ± SD | | | |
|---|---|---|---|---|---|
| Modality | Model | 3 months | 6 months | 9 months | 12 months |
| Blood markers (*n* = 231) | SimTA | **0.83 ± 0.03** | 0.80 ± 0.01 | 0.80 ± 0.02 | 0.80 ± 0.04 |
| | TSimTA | 0.82 ± 0.03 | **0.81 ± 0.02** | **0.82 ± 0.03** | **0.81 ± 0.03** |
| Imaging features (organ volumes) (*n* = 111) | SimTA | 0.63 ± 0.07[a] | 0.59 ± 0.05[a] | 0.58 ± 0.04[a] | 0.57 ± 0.03[b] |
| | TSimTA | **0.65 ± 0.07**[a] | **0.62 ± 0.05**[a] | **0.59 ± 0.03**[a] | 0.57 ± 0.01[b] |
| Medications (*n* = 231) | SimTA | 0.68 ± 0.03 | **0.68 ± 0.07** | 0.67 ± 0.05 | 0.66 ± 0.03 |
| | TSimTA | **0.69 ± 0.02** | 0.66 ± 0.04 | **0.68 ± 0.02** | **0.68 ± 0.02** |

Abbreviations: AUC: area under the receiver operating characteristic curve; SimTA: simple temporal attention[31]; TSimTA: transformer-based SimTA. All results are reported across 3 folds of cross-validation. SimTA contained 3 layers, while TSimTA contained 3 SimTA layers embedded in 1 transformer block. *P*-values for each fold were significant in all experiments except where marked with [a] for non-significance (n.s.) in 2 folds and [b] for n.s. in 3 folds. The best AUC results per modality and endpoint are highlighted in bold.

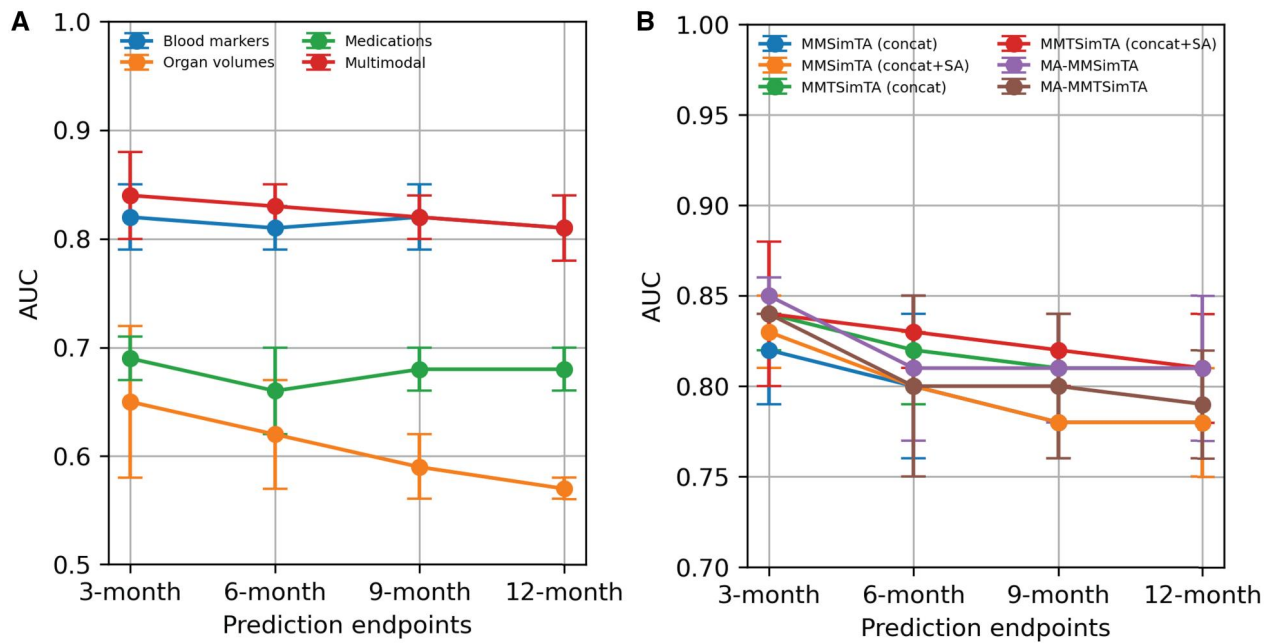

**Figure 2.** (A) The performance of the extended unimodal modality-specific, TSimTA, and the multimodal best-performing, MMTSimTA$_{concat+SA}$, classifiers. (B) The performance of the different variants of the integration methods, using MMSimTA[31], MMTSimTA and late fusion-based integration methods. Results in A and B are reported using data spanning between 3 months pretreatment and 3 months after.

**Table 3.** Comparative analysis of the performance of different multimodal classifiers, using up to 3-month on-treatment data (blood markers, volumes of organs, and medications).

| | | | Mean AUC ± SD | | | |
|---|---|---|---|---|---|---|
| Included patients | Model | Experiment | 3 months | 6 months | 9 months | 12 months |
| All patients with missing modalities (*n* = 231) | MMSimTA | Concat[a] | 0.82 ± 0.03 | 0.80 ± 0.04 | 0.78 ± 0.02 | 0.78 ± 0.03 |
| | | Concat + SA | 0.83 ± 0.02 | 0.80 ± 0.03 | 0.78 ± 0.02 | 0.78 ± 0.03 |
| | MA-MMSimTA | Mean | **0.85 ± 0.01** | 0.81 ± 0.04 | 0.81 ± 0.03 | **0.81 ± 0.04** |
| | MMTSimTA | Concat | 0.84 ± 0.02 | 0.82 ± 0.03 | 0.81 ± 0.03 | **0.81 ± 0.03** |
| | | Concat + SA | 0.84 ± 0.04 | **0.83 ± 0.02** | **0.82 ± 0.02** | **0.81 ± 0.03** |
| | MA-MMTSimTA | Mean | 0.84 ± 0.005 | 0.80 ± 0.05 | 0.80 ± 0.04 | 0.79 ± 0.03 |

Abbreviations: AUC: area under the receiver operating characteristic curve; Concat: aggregation via concatenation; MA-MMSimTA: model-agnostic multimodal integration based on the prediction probabilities of the unimodal SimTA; MA-MMTSimTA: model-agnostic multimodal integration based on the prediction probabilities of the unimodal TSimTA; MMSimTA: multimodal SimTA; MMTSimTA: multimodal TSimTA; SA: self-attention in a transformer block; SimTA: simple temporal attention; TSimTA: transformer-based SimTA. All results are reported across 3 folds of cross-validation. SimTA contained 3 layers, while TSimTA contained 3 SimTA layers embedded in one transformer block. *P*-values of each fold were significant in all experiments. The best AUC results per multimodal model and endpoint are highlighted in bold.

[a] Baseline MMSimTA model without positional encoding, which is a variant of the model proposed in Ref.[31]

**Table 4.** Comparative analysis of the performance of the best-performing multimodal models using up to 3- and 6-month on-treatment data of the same subsets of patients

| Included patients | Base model | Experiment | Mean AUC ± SD | | | |
|---|---|---|---|---|---|---|
| | | | 3 months | 6 months | 9 months | 12 months |
| 6 months (n = 184) | MMSimTA | Concat + SA | 0.75 ± 0.09 | 0.77 ± 0.05 | 0.76 ± 0.04 | 0.75 ± 0.03 |
| | MA-MMSimTA | Mean | 0.73 ± 0.09 | 0.77 ± 0.04 | 0.75 ± 0.04 | 0.76 ± 0.04 |
| | MMTSimTA | Concat + SA | **0.81 ± 0.07** | **0.82 ± 0.03** | **0.80 ± 0.02** | **0.79 ± 0.01** |
| | MA-MMTSimTA | Mean | 0.75 ± 0.06 | 0.77 ± 0.05 | 0.76 ± 0.04 | 0.76 ± 0.01 |
| 3 months (n = 184) | MMTSimTA | Concat + SA | (No death) | 0.76 ± 0.06 | 0.75 ± 0.03 | 0.75 ± 0.04 |

Abbreviations: AUC: area under the receiver operating characteristic curve; Concat: aggregation via concatenation; MA-MMSimTA: model-agnostic multimodal integration based on the prediction probabilities of the unimodal SimTA; MA-MMTSimTA: model-agnostic multimodal integration based on the prediction probabilities of the unimodal TSimTA; MMSimTA: multimodal SimTA[31]; MMTSimTA: multimodal TSimTA; SA: self-attention in a transformer block; SimTA: simple temporal attention[31]; TSimTA: transformer-based SimTA. All results were reported across 3 folds of cross-validation. SimTA contained 3 layers, while TSimTA contained 3 SimTA layers embedded in one transformer block. P-values of each fold were significant in all experiments. The best AUC results are highlighted in bold.

12-month survival, respectively). While comparing the extended MA-MMTSimTA to the best unimodal TSimTA model of blood data, integrating modalities via late fusion did not always perform better than the best unimodal modality (AUCs$_{MA-MMTSimTA}$ of 0.84, 0.80, 0.80, 0.79 vs AUCs$_{T-SimTA}$ of 0.82, 0.81, 0.82, 0.81; DeLong P-values of .02, >.05, >.05, >.05 for the prediction of 3-, 6-, 9-, and 12-month survival, respectively). Table 3 summarizes the results of all variants of multimodal experiments, while Table S8 displays the DeLong P-values of different pairs of experiments. Figure 2B illustrates the prognostic performance of the different multimodal experiments.

### Longer-term on-treatment data shows stronger prognostic performance

We evaluated the performance of the best-performing baseline and extended transformer-based multimodal models, with both intermediate and late fusion, using longer-term longitudinal on-treatment data (spanning between 3 months pretreatment to 6 months after, instead of 3 months after). To compare with 3-month survival, we used the best-performing model variant, MMTSimTA$_{concat+SA}$, and the same subsets of patients (n = 184) with long-term (6-month) on-treatment data by using only their corresponding short-term (3-month) on-treatment data. Predictive performance was better using longer-term on-treatment data (AUCs$_{6-month}$ of 0.81, 0.82, 0.80, 0.79 for 3-, 6-, 9-, and 12-month survival prediction vs AUCs$_{3-month}$ 0.76, 0.75, 0.75 for 6-, 9-, and 12-month survival; DeLong P-values of >.05, .01, .04 for the matched endpoints of 3- to 6-, 6- to 9-, and 9- to 12-month survival, where 3- to 6-month survival refers to the 3-month survival with the inclusion of the 6-month FU data, and the 6-month survival with the inclusion of the 3-month FU data), as shown in Table 4.

### Discussion

This study aimed to develop a novel neural network architecture by harnessing existing successful AI modules for the prediction of overall survival (mortality) in immunotherapy using noninvasive longitudinal and multimodal medical data. We predicted mortality at specific future endpoints: 3, 6, 9, and 12 months. We included longitudinal blood test measurements, volumes of organs extracted from contrast-enhanced CT imaging, and medications administered to patients over time. Performance was evaluated using early (3-month) and late (6-month) on-treatment data (Tables 3 and 4). We extended the transformer encoder block[43] with the SimTA module,[31] resulting in a transformer-based SimTA model and its multimodal variant (TSimTA and MMTSimTA, respectively).

Blood data showed the best unimodal prognostic capacity across all prediction endpoints. The extended TSimTA model demonstrated mixed effects on the performance, in terms of AUCs, in comparison to the SimTA model (Table 2 and Table S8). Overall, multimodal models demonstrated mild improvements in prognostic performance over unimodal models, which could be attributed to the dominant performance of the best unimodal modality (blood data) over all included modalities. The extended transformer-based multimodal model, MMTSimTA, demonstrated stronger prognostic performance than the baseline multimodal model, MMSimTA (Table 3 and Table S8). The additional layers with non-linearity and skip connection in the transformer-based model likely enhanced the model's ability to learn from more complex multimodal data. We compared our jointly trained multimodal models incorporating intermediate fusion to a model-agnostic late fusion method. We observed better performance with intermediate fusion via the extended transformer-based model, while late fusion performed better with the baseline models. Our late fusion approach is simpler compared to the intermediate fusion approach. It also assigns equal weights to unimodal modalities, regardless of their individual performance quality. Therefore, it is likely that more complex unimodal models, the extended transformer-based models, may benefit from a more sophisticated integration method like intermediate (feature-level) fusion, which allows better feature-level interactions between modalities, especially in the presence of dominant modalities. Conversely, simpler unimodal models, using the baseline SimTA network, may not necessarily benefit from feature-level interactions when dominant modalities are present, and may therefore perform better with simpler integration methods, such as late (decision-level) fusion. Further empirical research is needed to confirm this finding. Overall, the extended transformer-based multimodal model, specifically the MMTSimTA$_{concat+SA}$ variant, demonstrated the best performance in our study (Table 3), suggesting that intermediate fusion with a more sophisticated unimodal feature extractor may better address the requirements of multimodal data.

In addition to showing the potential of AI models to predict survival using early-stage on-treatment data, we evaluated the prognostic performance using patients with longer-term, 6-month, on-treatment data. The models utilizing longer-term on-treatment data performed better in all prediction endpoints compared to the models utilizing the corresponding 3-month on-treatment data (Table 4). This could be possibly due to the fact that the models focused on predicting mortality based on more recent examinations, which were acquired after longer exposure to treatment.

Several previous studies utilized longitudinal data of a single modality in their analysis, such as CT imaging in the study of Xu et al[52] for the prediction of lung cancer treatment response. They combined pretrained convolutional neural network (CNN) encoders and recurrent layers with gated recurrent units to encode longitudinal imaging. Similarly, a distanced long short-term memory network along with time-distanced gates was proposed by Gao et al[53] to model longitudinal CT scans for lung cancer diagnosis. Subsequently, a time-distance vision transformer, with positional encodings and scaled SA module, was proposed in the study of Li et al[54] for the same prediction task. Another longitudinal image analysis method was proposed in the study of Zhao et al,[55] which combined CNN encoders with the SimTA module[31] to summarize longitudinal CT scans for lung adenocarcinoma invasiveness prediction. Our study also followed a similar approach of extending the SimTA module within the state-of-the-art transformer architecture for survival prediction of immunotherapy-treated cancer patients. However, we utilized multimodal data instead of only one modality to harness the potentially complementary information of different modalities. Similar to our study, the MIA-Prognosis framework, which previously introduced the SimTA module in 2 subsequent studies,[31,33] predicted the response of immunotherapy utilizing longitudinal and multimodal data (CT radiomics, blood, and baseline clinical data) in advanced-stage NSCLC patients. They integrated the concatenated modality vectors via an MLP in an end-to-end trained multimodal model. Another study by Farina et al utilized longitudinal blood, CT scans, and clinical data to predict progression-free survival of NSCLC patients.[39] They utilized concatenated modality-based longitudinal features to train modality-specific random forest models, which were subsequently combined using a model-agnostic late fusion method. Overall survival of immunotherapy patients, with a smaller cohort, different machine learning models and model-agnostic late fusion multimodal integration method, was predicted in our previous study.[40] A similar dominant performance of blood markers as an individual modality was also observed, as well as a modest improvement in the prognostic performance of combined blood data, CT imaging, and clinical parameters.

In the context of this study, specific limitations may impact the generalizability of our findings. Firstly, we tested our method using only specific data modalities of immunotherapy-treated cancer patients, with volumes of organs representing the included imaging features. Such features might not be as prognostic as learned CT imaging or tumor-specific features. We also excluded a large number of patients ($n = 280$) to ensure the availability of a minimum of 3 months of FU data for assessing our longitudinal method. This exclusion criterion may have introduced selection bias in our cohort, as patients with shorter FU durations were not considered. We assessed the significance of differences between the included and excluded cohorts for several variables, and some of these variables showed significant differences (Table S2). To mitigate this, future studies could consider a more flexible FU criterion. Our analysis was also based on data from a single cancer center. Future external validation is essential to assess the robustness of our results across different populations and clinical settings. Furthermore, the utilized temporal attention module is designed to prioritize the most recent examinations (Figure S3), which could be relevant in healthcare. However, allowing algorithms to learn to attend to key examinations in longitudinal data could also be a promising domain. Other methods that can efficiently model the dynamics of longitudinal data in healthcare, including neural differential equations, selective state space models, and dynamic Bayesian networks, also present promising avenues for precision oncology and the development of digital twin models when integrated with multimodal data.[56] Another limitation could be the utilization of multimodal dropout, while using it allows for easy application of multimodal learning with incomplete modalities, it introduces additional unwanted noise to the learned representations, which, in turn, could affect the learning process. An alternative approach would be to generate missing modalities based on patient similarities, with a variant of it being introduced in the $M^3$Care platform by Zhang et al.[57]

Finally, in concluding this study, 2 potential directions sound viable for future multimodal research. The first is to use only the dominant and most informative modalities, which can potentially predict the studied endpoint, in our case, blood data for the prediction of survival, avoiding the additional costs of integrating more modalities with limited to no added value. Further research on potential modality collapse and/or competition[58] is also required in this context for medical applications. The other direction is to include all available modalities of patients, which could indeed be challenging both cost-wise and methodologically, given the heterogeneity and the high dimensionality of the different medical data sources. The latter requires the availability of pretrained medical (foundation) models with the capability to accurately summarize high-dimensional data into feature vectors to facilitate efficient multimodal learning research. Despite these challenges, multimodal models trained with all modalities could be more generalizable and adaptable to other tasks, as each medical modality may contribute differently depending on the prediction task. Lastly, and most importantly, a crucial requirement for successful multimodal research in healthcare is the availability of large-scale multimodal medical datasets and the development of advanced AI methods, which can only be achieved through improved collaborations between clinicians of different specialties and AI experts.

## Acknowledgments

Research at the Netherlands Cancer Institute is supported by institutional grants from the Dutch Cancer Society and the Dutch Ministry of Health, Welfare and Sport. This work was also supported by the Radboud Healthy Data Program. The computational infrastructure used in this project was made possible via generous support from the Maurits en Anna de Kock Stichting (toekenning: 2019-8) and the NVIDIA Academic GPU Program. We would also like to acknowledge the Research High Performance Computing (RHPC) facility of the Netherlands Cancer Institute—AVL Hospital.

## Author contributions

Melda Yeghaian: Conceptualization, Data curation, Formal analysis, Investigation, Methodology, Software, Validation, Visualization, Writing—original draft, Writing—review & editing. Zuhir Bodalal: Conceptualization, Writing—review & editing. Daan van Den Broek: Conceptualization, Resources. John B.A.G. Haanen: Conceptualization, Resources. Regina G.H. Beets-Tan: Conceptualization, Funding acquisition, Resources. Stefano Trebeschi: Conceptualization, Data curation, Methodology, Supervision, Writing—review & editing. Marcel A.J. van Gerven: Funding acquisition, Supervision, Writing—review & editing.

## Supplementary material

Supplementary material is available at *Journal of the American Medical Informatics Association* online.

## Funding

This study was partially funded by the Radboud Healthy Data consortium.

## Conflicts of interest

The authors declare no competing interests.

## Data availability

The code will be publicly accessible on GitHub at https://github.com/MeldaYeghaian/MMTSimTANet.